\documentclass{bmvc2k}

\usepackage{amsmath}
\usepackage{amssymb}

\usepackage{tabulary}
\newcolumntype{K}[1]{>{\centering\arraybackslash}p{#1}}

\mathchardef\mhyphen="2D
\title{S$^3$D: Single Shot multi-Span Detector via Fully 3D Convolutional Network}

\addauthor{Da Zhang}{dazhang@cs.ucsb.edu}{1}
\addauthor{Xiyang Dai}{xdai@umiacs.umd.edu}{2}
\addauthor{Xin Wang}{xwang@cs.ucsb.edu}{1}
\addauthor{Yuan-Fang Wang}{yfwang@cs.ucsb.edu}{1}

\addinstitution{
 Computer Vision Lab\\
 Dept. of Computer Science\\
 UC Santa Barbara, USA
}
\addinstitution{
Dept. of Computer Science\\
University of Maryland, USA
}

\runninghead{ZHANG ET AL.}{S$^3$D: Single Shot multi-Span Detector}

\def\eg{\emph{e.g}\bmvaOneDot}

\def\etal{\emph{et al}\bmvaOneDot}

\begin{document}

\maketitle


\begin{abstract}
	In this paper, we present a novel Single Shot multi-Span Detector for temporal activity detection in long, untrimmed videos using a simple end-to-end fully three-dimensional convolutional (Conv3D) network. Our architecture, named S$^3$D, encodes the entire video stream and discretizes the output space of temporal activity spans into a set of default spans over different temporal locations and scales. At prediction time, S$^3$D predicts scores for the presence of activity categories in each default span and produces temporal adjustments relative to the span location to predict the precise activity duration. Unlike many state-of-the-art systems that require a separate proposal and classification stage, our S$^3$D is intrinsically simple and dedicatedly designed for single-shot, end-to-end temporal activity detection. When evaluating on THUMOS'14 detection benchmark, S$^3$D achieves state-of-the-art performance and is very efficient and can operate at 1271 FPS. 
\end{abstract}


\section{Introduction}
  Advances in deep Convolutional Neural Network (CNN) have led to significant progress in video analysis over the past few years. While the performance of activity recognition has improved a lot~\cite{wang2011action,wang2013action,simonyan2014two,feichtenhofer2016convolutional,wang2015towards,tran2015learning}, the detection performance still remains unsatisfactory~\cite{wang2014action,yeung2016end,shou2016temporal}. Comparing to activity recognition, which only aims at classifying the categories of manually trimmed video clips, activity detection is for detecting and recognizing activity instances from long, untrimmed video streams. It is substantially more challenging, as it is expected to handle activities with variable lengths, predicting both the activity category and the precise temporal boundaries of each instance.
  
  A typical framework used by many state-of-the-art systems~\cite{oneata2014lear,shou2016temporal,wang2014action,cdc_shou_cvpr17} is \textit{detection by classification}, where temporal proposals are generated by sliding windows~\cite{oneata2014lear,shou2016temporal} or advanced proposal methods~\cite{wang2016actionness,caba2016fast} and separate activity classifier is applied to predict the final detection results. However, there may be certain limitations to these frameworks: (1) Temporal proposal and classification are independent processes and optimized separately with different networks, resulting in sub-optimal performance, (2) the classification network only takes the proposal frames as input, thus forbidding it to see a larger temporal context which can be beneficial, and (3) this two-stage approach is usually slow due to inefficient proposal method and duplicate operations repeated in the proposal and classification stages.
    
\begin{figure*}
\begin{center}
\includegraphics[width=1.0\linewidth]{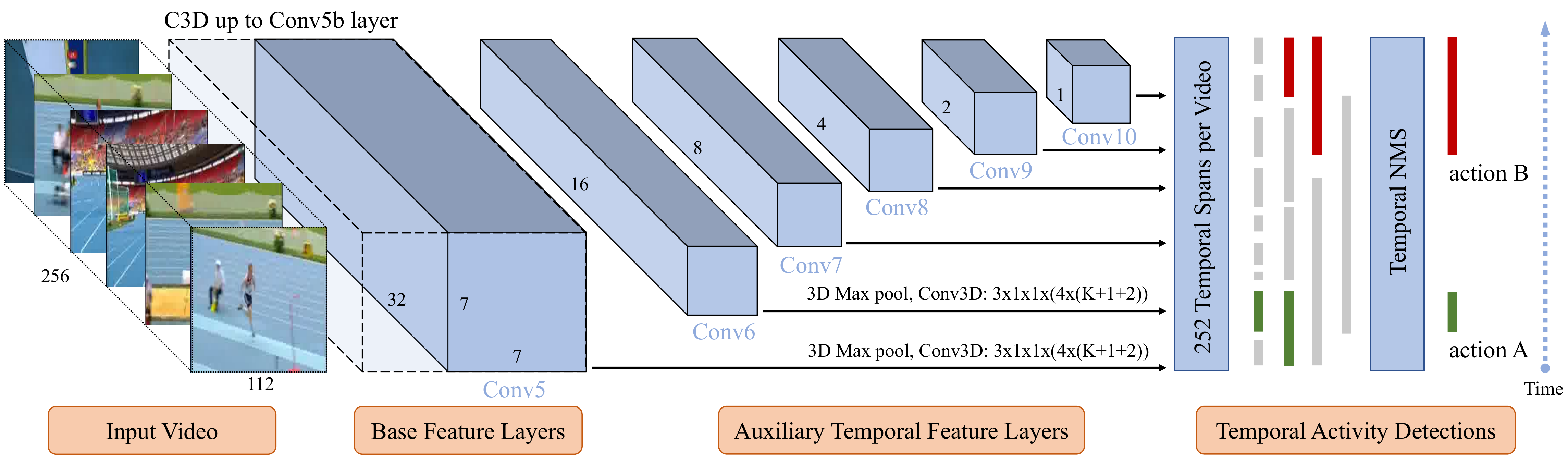}
\end{center}
   \caption{S$^3$D network architecture: Our network takes a video of $256$ frames with spatial size $112\times 112$ as input and computes base features using a standard C3D~\cite{tran2015learning} network up to $\mathsf{conv5b}$. We add auxiliary Conv3D layers on top of $\mathsf{conv5}$ to produce a temporal feature hierarchy with multi-scale default spans at each layer. For each temporal feature map cell, we predict $K$ class confidence scores, $1$ activity confidence score and $2$ location offsets with a set of Conv3D filters. Temporal NMS is applied to produce the final detection results. Refer to Figure~\ref{fig:anchor} for a detailed illustration of default spans.}
\label{fig:s3d}
\end{figure*}

    We propose a \textit{Single Shot multi-Span Detector (S$^3$D)}, a simple yet novel fully Conv3D-based framework for activity detection in continuous untrimmed video streams. As illustrated in Figure~\ref{fig:s3d}, S$^3$D produces a fixed-size collection of temporal spans and scores for the presence of activity class instances in those spans, followed by a temporal non-maximum suppression step to generate the final detection results. S$^3$D is a highly-unified network by eliminating explicit temporal proposal and classification stages and solving the detection problem in one single shot. We set multi-scale default spans at feature maps with different temporal resolutions to naturally handle activities of different lengths. Furthermore, we predict the temporal offsets to adjust each default span in order to predict precise temporal boundaries. The network takes as input a whole video stream, allowing our scheme to see a larger temporal context and produce better detection results. The whole network is end-to-end trainable with a joint loss to directly maximize the detection performance.

	The contributions of our paper are: (1) We introduce S$^3$D, a single shot end-to-end activity detection model based completely on  Conv3D networks that can effectively predict both the precise temporal boundaries and confidence scores of multiple activity categories in untrimmed videos. (2) We demonstrate experimentally that our S$^3$D achieves state-of-the-art performance on temporal activity detection task on THUMOS'14 benchmark. (3) Besides its strong performance, the simple S$^3$D network is also very efficient and can run at 1271 FPS on a single GPU. Our code is available at \href{https://github.com/dazhang-cv/S3D}{https://github.com/dazhang-cv/S3D}.

\section{Related Work}
Here, we review relevant works in activity recognition, object detection, and temporal activity detection. Other works on spatial-temporal activity detection and temporal video segmentation are beyond the scope of this paper.

\noindent \textbf{Activity Recognition.} Activity recognition is an important research topic for video analysis and has been extensively studied in the past few years. Earlier methods were often based on hand-crafted visual features. Approaches include improved Dense Trajectory (iDT)~\cite{wang2011action,wang2013action}, feature encoding with Fisher Vector (FV)~\cite{perronnin2010improving,oneata2013action}, VLAD~\cite{jegou2010aggregating}, etc. With the vast successes of deep learning methods~\cite{zhang2017deep,8100129,chen2017coherent,hmm2018deep, chen2017stylebank,he2016deep,simonyan2014very}, recent works, such as two-stream networks~\cite{simonyan2014two,feichtenhofer2016convolutional,wang2015towards}, 3D CNN architecture (C3D)~\cite{tran2015learning} and I-3D~\cite{carreira2017quo}, adopted deep neural networks and significantly improved the performance. However, most methods assume well-trimmed videos, where the activity of interests lasts for the entire clip duration. Although they do not consider the difficult task of localizing activity instances, these methods are widely used as the base network for the detection task.

\noindent \textbf{Object Detection.} Activity detection in untrimmed videos is closely related to object detection~\cite{girshick2014rich,ren2015faster,liu2016ssd} in spatial images, where detection is performed by classifying region proposals into foreground classes or a background class. Earlier work~\cite{girshick2014rich} relied on an external region proposal method and trained a CNN classifier to classify each proposed region. Faster-RCNN~\cite{ren2015faster} incorporated a region proposal network and RoI pooling to jointly generate and classify region proposals with a single network, resulting in a large improvement of the accuracy and efficiency. SSD~\cite{liu2016ssd} completely eliminated proposal generation and subsequent feature re-sampling stages and encapsulated all computation in a single network to directly output object locations and confidence scores. Our network is inspired by SSD~\cite{liu2016ssd} and adopt similar design philosophies into temporal activity detection. Like SSD~\cite{liu2016ssd}, our S$^3$D model is also designed for both accuracy and efficiency.

\noindent \textbf{Temporal Activity Detection.} Unlike activity recognition, the detection task focuses on learning how to detect activity instances in untrimmed videos with annotated temporal boundaries and instance category. The problem has recently received significant research attention due to its potential application in video data analysis. 

Previous works on activity detection mainly used sliding windows as candidates and classified them with activity classifiers trained on multiple features~\cite{oneata2013action,gaidon2013temporal,jain2014action,mettes2015bag,tang2013combining}. Recently, some approaches  bypassed the need for exhaustive sliding window search by proposing better temporal proposal schemes~\cite{caba2016fast,escorcia2016daps,van2015apt,mettes2016spot,yu2015fast,buch2017sst}. Along this line of attack, some recent works incorporated deep networks into the detection framework and  obtained improved performance~\cite{shou2016temporal,cdc_shou_cvpr17,Xu2017iccv,dai2017temporal,zhao2017temporal}: S-CNN~\cite{shou2016temporal} proposed a multi-stage CNN which adopted 3D ConvNet with multi-scale sliding window to boost accuracy; CDC~\cite{cdc_shou_cvpr17} used temporal deconvolutional network to generate per-frame classification scores for refining temporal boundaries; R-C3D~\cite{Xu2017iccv} proposed an end-to-end trainable activity detector based on Faster-RCNN~\cite{ren2015faster}; Dai \etal ~\cite{dai2017temporal} explicitly modeled temporal contextual information into the proposal stage; SSN~\cite{zhao2017temporal} utilized temporal pyramid pooling to model the complicated temporal structures and achieved state-of-the-art performance. However, all these methods require a separate temporal proposal and activity classification method. 

Most recently, several attempts were made towards single shot temporal activity detection: SSAD~\cite{DBLP:conf/mm/LinZS17} proposed to directly predict activity instances in untrimmed videos with a separate feature extraction and detection network. SS-TAD~\cite{buch2017end} have investigated the use of gated recurrent memory module in a single-stream detection framework. Our approach is one of the first within this group to propose a highly-integrated detection architecture. With a simple end-to-end Conv3D network which learns directly from raw video frames to final detection outputs, S$^3$D is able to jointly optimize feature representation and prediction layers, resulting in a simple, fast and robust architecture achieving both state-of-the-art performance and fast runtime speed. 

\section{Approach}
	We introduce a \textit{Single Shot multi-Span Detector (S$^3$D)}, a simple yet novel fully Conv3D-based framework for activity detection in long untrimmed video streams. The S$^3$D approach, illustrated in Figure~\ref{fig:s3d}, is based on a feed-forward fully Conv3D network that produces a fixed-size collection of temporal spans and scores for the presence of activity class instances in those spans, followed by a temporal NMS step to generate the final detection results.
    
\subsection{Model}
    
    Our model consists of four major components: base feature layers, auxiliary temporal feature layers, multi-scale default spans and convolutional predictors. The \textit{base feature layers} are used to extract high-level features given an input video stream. We then add \textit{auxiliary temporal feature layers} to generate rich spatial-temporal feature hierarchies. These layers decrease in temporal dimension progressively and allow predictions of temporal spans at different locations and scales. We associate \textit{multi-scale default spans} with each feature map cell and the default spans tile the feature map in a convolutional manner. At each feature map cell, we predict the temporal offsets relative to the default span in the cell, as well as the confidence scores that indicate the presence of an activity instance in each of those spans. These are done by adding \textit{convolutional predictors} on top of each cell. 

	\textbf{Base Feature Layers.} We use Conv3D filters to extract rich feature hierarchies from a given input video stream. Specifically, the input to our model is a sequence of RGB video frames which can be represented as a tensor with dimension $\mathbb{R}^{L\times H\times W\times 3}$, where $L$ is the number of frames, $H$ and $W$ are the height and width of each frame. We apply the standard C3D architecture~\cite{tran2015learning} as it has been proven as an effective building block in prior works~\cite{Xu2017iccv,DBLP:conf/mm/LinZS17,buch2017end}. We adopt the Conv3D layers ($\mathsf{conv1a}$ to $\mathsf{conv5b}$) of C3D and generate a feature map $C_{conv5}\in \mathbb{R}^{\frac{L}{8}\times \frac{H}{16}\times \frac{W}{16}\times 512}$. We use $C_{conv5}$ as our base feature since it is a rich yet compact spatial-temporal representation of the input video stream. 

\textbf{Auxiliary Temporal Feature Layers.} To allow the model to predict variable scale temporal spans, we add temporal feature layers to the end of the base feature layers. Similar to~\cite{tran2015learning}, we first down sample $C_{conv5}$ by a factor of $2$ in both spatial and temporal dimension via 3D max pooling and then add auxiliary Conv3D layers to produce a sequence of feature maps that progressively decrease in temporal dimension while keeping the same spatial resolution. In more detail, we stack Conv3D layers with temporal kernel size $3$ to extend the temporal receptive field and the stride is set to $2$ for progressively decreasing the temporal dimension. We also add bottleneck Conv3D layers to help prevent over-fitting and improve runtime efficiency. The detailed network configurations are illustrated in Figure~\ref{fig:s3d} when $L=256$ and $H=W=112$. 
    
    The network is intrinsically simple by only applying Conv3D filters, but builds a rich feature hierarchy by summarizing a continuous video stream in multiple temporal resolutions, allowing us to add default temporal spans at certain layers to get temporal predictions at multiple scales.
    
\textbf{Multi-scale Default Spans.} To handle different activity locations and scales, ~\cite{shou2016temporal} suggests processing the video at different segment levels and combining the results afterward, while~\cite{buch2017end} uses a gated recurrent network to assign a number of anchors at different time steps. However, by utilizing feature maps from several different layers in a single network for prediction we can mimic the same effect, while also sharing parameters across all temporal scales. We use feature maps with different temporal resolutions for detection since earlier feature maps have higher resolution and capture finer details of the input video, and deeper feature maps have larger receptive fields and contain more temporal contexts. 

   In our design, we use $\mathsf{conv5}$ to $\mathsf{conv10}$ as our temporal feature maps and associate a set of multi-scale default spans with each temporal feature map cell. We design the tiling of default spans so that specific feature maps learn to be responsive to particular locations and lengths of the activities. Regrading a temporal feature map $f$ with temporal length $L_{f}$, the scale of the default spans for this feature map is set as $S_{f}=\frac{1}{L_{f}}$ (as the input video length is normalized to $1$). We impose different scale ratios for the default spans, and denote them as $r\in \{0.25, 0.5, 0.75, 1.0\}$. We can compute the length ($l_{f}^{r} = S_{f}\cdot r$) for each default span, and we set the center of each default span to $\frac{i+0.5}{L_{f}}$, where $i$ indicates the $i\mhyphen{th}$ temporal feature cell, $i\in [0,L_{f})$. So for an temporal feature map with length $L_{f}$ and $R$ different scale ratios ($R=4$), the number of default spans is $L_{f}\cdot R$. 
    
\begin{figure*}
\begin{center}
\includegraphics[width=1.0\linewidth]{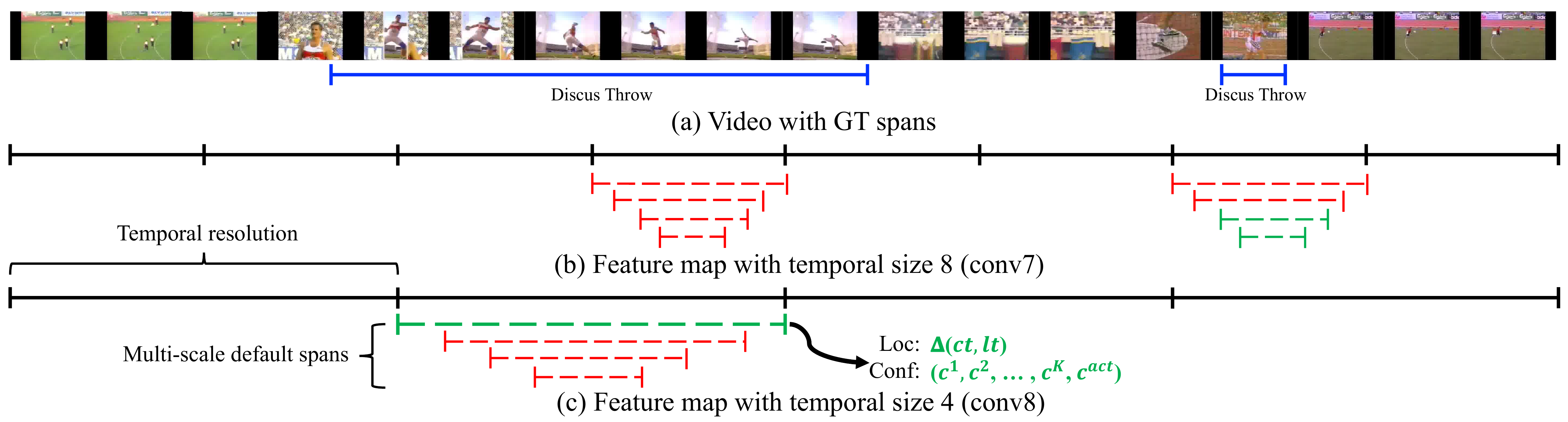}
\end{center}
   \caption{S$^3$D framework. (a) Input video with temporal ground-truth annotations. We evaluate a small set (\eg $4$) of multi-scale default spans at each location in several feature maps with different temporal resolutions (\eg  $\mathsf{conv7}$ in (b) and $\mathsf{conv8}$ in (c)). For each default span, we predict both the temporal offsets and the confidences for presence of activity and all activity categories. At training time, we match the default spans to the ground truth spans.} 
\label{fig:anchor}
\end{figure*}

    By combining predictions for all default spans with different scales from all locations of multi-scale feature maps, we have a diverse set of predictions, covering various activity locations and lengths. A concrete example is illustrated in Figure~\ref{fig:anchor} where $L_{conv7}=8$ and $L_{conv8}=4$ for feature map $\mathsf{conv7}$ and $\mathsf{conv8}$ respectively. 
    
\textbf{Convolutional Predictors.} Each temporal feature layer can produce a fixed set of detection predictions using a set of Conv3D filters. These are indicated on top of the feature network architecture in Figure~\ref{fig:s3d}. For a temporal feature map $C_{f}\in \mathbb{R}^{L_{f}\times H_{f}\times W_{f}\times d_{f}}$, the basic operation for predicting parameters of a potential temporal detection is a $3\times H_{f}\times W_{f}$ kernel that produces scores for activity presence and categories, or temporal offsets relative to the default location and scale. Specifically, for each default span at a given temporal location, we compute $K$ positive class confidence scores plus one activity confidence score and two temporal offsets. This results in a total of $(K+1+2)\times R$ filters that are applied around each location in the feature map, yielding $(K+1+2)\times R\times L_{f}$ outputs for a temporal feature map $C_{f}$. For an illustration of default spans, please refer to Figure~\ref{fig:anchor}. Each default span gets a prediction score vector $v_{pred}=(c^{1}, c^{2}, ... , c^{K}, c^{act}, \Delta ct, \Delta lt)$ with length $K+1+2$, where $c^{act}$ is a class-agnostic confidence score to estimate the presence of activity, $c^{1}$ to $c^{K}$ are used to predict default span's category and $\Delta ct, \Delta lt$ are temporal offsets relative to the locations of default spans.

\subsection{Training}
\label{subsec:train}
	The key step of training S$^3$D is that the ground truth information needs to be assigned to specific outputs in the fixed set of detector outputs. Once this assignment is determined, the loss function and back propagation are applied. We also discuss training data construction and hard negative mining strategies used in our model.
    
\textbf{Training Data Construction and Augmentation.} In theory, because S$^3$D is a fully Conv3D network, it can be applied to an input of arbitrary size. Therefore, our S$^3$D network can operate on videos of variable lengths. In practice, due to GPU memory limitations, we slide a temporal window of size $L$ frames on the video and feed each windowed segment individually into the S$^3$D network to obtain temporal detections. Although the input window size is fixed, we decode the input video stream with a small frame rate, allowing the network to encode enough temporal contexts for precisely detecting activity instances. Therefore, given a set of training videos, we obtain a training collection of windows with temporal activity annotations inside each windowed video segment. To make the model more robust to various activity locations and scales, we further improve the training dynamics by augmenting the training videos with temporal and spatial jittering~\cite{tran2015learning}.
    
\textbf{Matching Strategy.} During training, we need to determine which default spans correspond to a ground truth detection and train the network accordingly. Specifically, for each default span, we compute the Intersection-over-Union (IoU) score with all ground truth instances. If the highest IoU score is higher than $0.5$, we match the default span with the corresponding ground truth span and regard it as positive, otherwise negative. So a ground truth instance can match multiple default spans while a default span can only match one ground truth instance at most. This simplifies the learning problem, allowing the network to predict high scores for multiple overlapping default spans.

\textbf{Hard Negative Mining.} After the matching step, most of the default spans are negatives. This introduces a significant imbalance between the positive and negative training examples. Instead of using all the negative examples, we sort them using the highest activity confidence loss for each default span and pick the top ones so that the ratio between the negatives and positives is nearly $1:1$. We found that this leads to faster optimization and a more stable training.

\textbf{Training Objective.} The training objective of S$^3$D is to solve a multi-task optimization problem. Let $x_{ij}^{k}=\{1,0\}$ be an indicator for matching the $i\mhyphen th$ default span to the $j\mhyphen th$ ground truth span of category $k\in [1,K]$, and $s_{i}$ be the highest IoU score with any ground truth spans. The overall objective loss function is a weighted sum of the localization loss (loc), class confidence loss (conf) and activity confidence loss (act):
\begin{equation}
	Loss = L_{loc}(x, t, g) + \alpha L_{conf}(x, c) + \beta L_{act}(s, c)
    \label{eq:trainobj}
\end{equation}
where $\alpha$ and $\beta$ are the weight terms used for balancing each part of the loss function.

The localization loss is a Smooth L1 loss~\cite{girshick2015fast} between the predicted temporal offsets ($t$) and the ground truth span parameters ($g$). In temporal domain, we regress to offsets for the center ($ct$) of the default span ($d$) and for its length ($lt$):
\begin{equation}
	L_{loc}(x, t, g) =\frac{1}{N_{pos}} \sum_{i}^{N_{pos}} \sum_{m\in \{ct, lt\}}x_{ij}^{k}smooth_{L1}(t_{i}^{m} - \hat{g}_{j}^{m})
    \label{eq:locloss}
\end{equation}
where $N_{pos}$ is the number of positive matching default spans in a batch, and the temporal offset parameters $\hat{g}_{j}^{m}$ are defined similarly like the bounding box offset in object detection~\cite{girshick2015fast}:
\begin{equation}
   \hat{g}_{j}^{ct} = \Delta ct_{i} = (g_{j}^{ct} - d_{i}^{ct})/d_{i}^{lt}\ \ \ \ \ \ \hat{g}_{j}^{lt} = \Delta lt_{i} = \log(\frac{g_{j}^{lt}}{d_{i}^{lt}})
    \label{eq:tempreg}
\end{equation}
where $g_{j}^{ct}$, $d_{i}^{ct}$ are the centers and $g_{j}^{lt}$, $d_{i}^{lt}$ are the lengths for the ground truth span and the matching default temporal span respectively.

The class confidence loss is a softmax loss over multiple class confidences ($c$):
\begin{equation}
	L_{conf}(x, c) = -\frac{1}{N_{pos}}\sum_{i}^{N_{pos}}x_{ij}^{k}\log(\hat{c}_{i}^{k})
    \label{eq:clsloss}
\end{equation}
where $\hat{c}_{i}^{k}=\frac{\exp(c_{i}^{k})}{\sum_{k}\exp(c_{i}^{k})}$ is the softmax probability for the ground truth class of this instance. The class confidence loss is only used to distinguish between multiple positive classes not including the background. We use another activity confidence score to predict activity class agnostic scores.

The activity confidence loss is a binary classification loss using sigmoid cross-entropy. Rather than using a hard ground truth score for positive ($1$) and negative ($0$), we use the IoU score $s_{i}$ as ground truth for each default span. This helps the training procedure since positive default spans are assigned different confidence levels based on its overlap with the ground truth span. We define the activity confidence loss as:
\begin{equation}
	L_{act}(s, c) = -\frac{1}{N}\sum_{i}^{N}(s_{i}\log({c}_{i}^{act}) + (1-s_{i})\log(1-{c}_{i}^{act}))
    \label{eq:clsloss}
\end{equation}
where $N$ is the number of total training default spans in a batch and $N=N_{pos}+N_{neg}$; ${c}_{i}^{act}$ is the predicted activity confidence score. Note that we separate the activity confidence score and class confidence scores via two separate losses. Comparing to only having one softmax classification loss containing all positive classes and one background class, we find this configuration is more robust, leads to better validation performance and makes the network architecture more flexible.

\subsection{Prediction}
\label{subsec:pred}
	Activity prediction in S$^3$D is single shot with one forward pass of the network. Given an input video stream, we generate all default spans with class confidence scores, activity confidence score and temporal location offsets. The temporal location offset is in the form of relative displacement of the center point and length of each instance as described in Equation~\ref{eq:tempreg}, which is applied on the default span to predict accurate start time and end time. Then the default spans with low activity confidence score will be filtered out and the remaining spans are refined via NMS with threshold value $0.5$. Each remaining span is considered as a positive prediction and assigned the activity label with the highest class confidence score, which we consider as the final temporal detection results of S$^3$D.

\section{Experiments}
\label{sec:exp}
	We evaluate the proposed framework on the THUMOS'14~\cite{jiang2014thumos} large-scale activity detection benchmark dataset. As shown in the experiments, our S$^3$D not only achieves state-of-the-art performance but also acquires fast runtime speed at 1271 FPS. 
    
\subsection{Experimental Setup}
\label{subsec:expsetting}
\noindent \textbf{Dataset}~\cite{jiang2014thumos}. The temporal activity detection task of THUMOS'14 dataset is challenging and widely used. Over 20 hours of video and 20 activity categories are involved and annotated temporally, resulting in 200 validation and 213 test untrimmed videos. Following the standard practice, we train our models on the validation set and evaluate them on the testing set. We follow the conventional metrics used in THUMOS'14, computing the Average Precision (AP) for each activity category and calculating mean AP (mAP) for evaluation.

\noindent \textbf{Implementation Details.} S$^3$D takes as input $L=256$ raw video frames with size $H=W=112$. We decode each video at 8 FPS and produce a collection of training windows. Thus, each window contains 32 seconds of a video stream and this is motivated by the fact that more than $99\%$ of activity instances in THUMOS'14 are less than 32 seconds. We use $\mathsf{conv5}$ to $\mathsf{conv10}$ as the temporal feature layers with temporal dimension $\{32, 16, 8, 4, 2, 1\}$ and associate a set of default spans at each temporal feature cell with four ratios $\{0.25, 0.5, 0.75, 1.0\}$, resulting in $252$ default spans in total; the default spans correspond to spans of duration between $0.25$s and $32$s uniformly distributed at different temporal locations. We initialize base feature layers with C3D weights pre-trained on Sports-1M by the authors in~\cite{tran2015learning}, and other layers from scratch. We allow all the layers of S$^3$D to be trained on THUMOS'14 with the end-to-end loss function. 

\subsection{Comparison with State-of-the-art}
\begin{table}
\begin{center}
\small
\begin{tabular}{K{2.5cm}|K{0.7cm} K{0.7cm} K{0.7cm} K{0.7cm} K{0.7cm}}
\hline
IoU threshold & 0.3 & 0.4 & 0.5 & 0.6 & 0.7 \\
\hline
S-CNN~\cite{shou2016temporal} & 36.3 & 28.7 & 19.0 & 10.3 & 5.3 \\
CDC~\cite{cdc_shou_cvpr17} & 40.1 & 29.4 & 23.3 & 13.1 & 7.9 \\
SSAD~\cite{DBLP:conf/mm/LinZS17} & 43.0 & 35.0 & 24.6 & - & - \\
TCN ~\cite{dai2017temporal} & - & 33.3 & 25.6 & 15.9 & 9.0 \\
R-C3D~\cite{Xu2017iccv} & 44.8 & 35.6 & 28.9 & - & - \\
SSN~\cite{zhao2017temporal} & \textbf{50.6} & 40.8 & 29.1 & - & - \\
SS-TAD~\cite{buch2017end} & 40.1 & - & 29.2 & - & 9.6 \\
\hline \hline
S$^3$D & 47.9 & \textbf{41.2} & \textbf{32.6} & \textbf{23.3} & \textbf{14.3} \\
\hline
\end{tabular}
\end{center}
\caption{Temporal activity detection mAP on THUMOS'14. The top performing methods in existing papers are shown. S$^3$D achieves state-of-the-art performance at different overlap threshold.  (- indicates that results are unavailable in the corresponding papers).}
\label{tb:THUMOS}
\end{table}  

The comparison results between our S$^3$D and other top-performing methods are summarized in Table~\ref{tb:THUMOS}, and our S$^3$D outperforms all previous state-of-the-art methods. Furthermore, S$^3$D improves the state-of-the-art by a large margin when the evaluation IoU thresholds are set at higher levels ($0.5$ to $0.7$), indicating its superior ability to predict precise temporal boundaries of different activities. 

In comparison with the proposed S$^3$D model: previous systems on top of C3D networks (S-CNN~\cite{shou2016temporal}, CDC~\cite{cdc_shou_cvpr17}) largely relies on good temporal proposals generated by external proposal methods, restricting them from directly optimizing the detection performance. R-C3D~\cite{Xu2017iccv} is able to process a long video stream and predict multi-scale activity instances, but it only applies anchors on a single feature map with fixed temporal dimension. With the proposed S$^3$D framework, we jointly optimize the feature representation and detection layers at different temporal levels by processing an untrimmed input video stream with enough temporal context.
  
\subsection{Ablation Study}

To understand S$^3$D better, we evaluate our network with different variants on THUMOS'14 to study their effects. For all experiments, we only change the certain part of the network and use the same evaluation settings. We compare the result of different variants using the mAP at IoU threshold $0.5$.
    
\begin{table}[h]
\begin{center}
\small
\begin{tabular}{K{2.5cm}|K{0.7cm} K{0.7cm} K{0.7cm} K{0.7cm}}
\hline
include $1.0$ span & \checkmark & \checkmark & \checkmark & \checkmark \\
include $0.25$ span  &  & \checkmark & \checkmark & \checkmark \\
include $0.5$ span  &  &  & \checkmark & \checkmark \\
include $0.75$ span  &  &  &  & \checkmark \\
\hline \hline
\# Spans & 63 & 126 & 189 & 252 \\
mAP@0.5 & 27.5 & 29.5 & 31.1 & \textbf{32.6} \\
\hline
\end{tabular}
\end{center}
\caption{Effects of various design choices on S$^3$D performance, the span with ratio 1.0 is included by default.}
\label{tb:design}
\end{table}
    
\noindent \textbf{Default Span Ratio.} By default, we use $4$ default spans per each temporal location. If we remove the spans with ratio $0.75$, the mAP drops by $1.5\%$. By further removing the spans with ratio $0.25$ and $0.5$, the mAP drops another $3.6\%$. By only keeping the span with ratio $1.0$, our model already has a strong performance (mAP $27.5\%$) since it already covers most ground truth instances in the dataset. Using a variety of default ratios make the task of predicting spans easier for the network and result in better performance.

\noindent \textbf{Span Regression.} The default spans are defined at fixed temporal locations. In order to generate precise predictions for starting and ending time of each activity instance, we adjust each default span by applying a temporal offset described in Equation~\ref{eq:tempreg}. This technique, which we call span regression, allows our model to predict temporal spans at much smaller granularities. As shown in Table~\ref{tb:layer}, span regression improves the mAP from $28.6\%$ to $32.6\%$.

\begin{table}[h]
\footnotesize
\begin{center}
\begin{tabular}{K{2.3cm} K{0.75cm} K{0.75cm} K{0.75cm} K{0.75cm} K{0.75cm} K{0.75cm}|K{1.2cm}|K{0.9cm}}
\hline
Span regression & conv5 & conv6 & conv7 & conv8 & conv9 & conv10 & mAP@0.5 & \# Spans \\
\hline
\checkmark & \checkmark & \checkmark & \checkmark & \checkmark & \checkmark & \checkmark & \textbf{32.6} & 252 \\
 & \checkmark & \checkmark & \checkmark & \checkmark & \checkmark & \checkmark & 28.6 & 252 \\
\checkmark & \checkmark & \checkmark & \checkmark & \checkmark & \checkmark &  & 31.8 & 248 \\
\checkmark & \checkmark & \checkmark & \checkmark & \checkmark &  &  & 30.7 & 240 \\
\checkmark & \checkmark & \checkmark & \checkmark &  &  &  & 27.6 & 224 \\
\hline
\end{tabular}
\end{center}
\caption{Effects of using multiple temporal feature layers and span regression.}
\label{tb:layer}
\end{table}

\noindent \textbf{Multi-scale Default Spans.} A major advantage of S$^3$D is using default spans of different scales on different temporal feature layers. To measure the advantage gained, we progressively remove layers and compare results. Table~\ref{tb:layer} shows a decrease in accuracy with fewer layers, dropping monotonically from $32.6\%$ to $27.6\%$. This is because that different layers are responsible for predicting temporal activities at different lengths, which reinforces the message that it is critical to spread spans of different scales over different layers.

\begin{figure*}
\begin{center}
\includegraphics[width=1.0\linewidth]{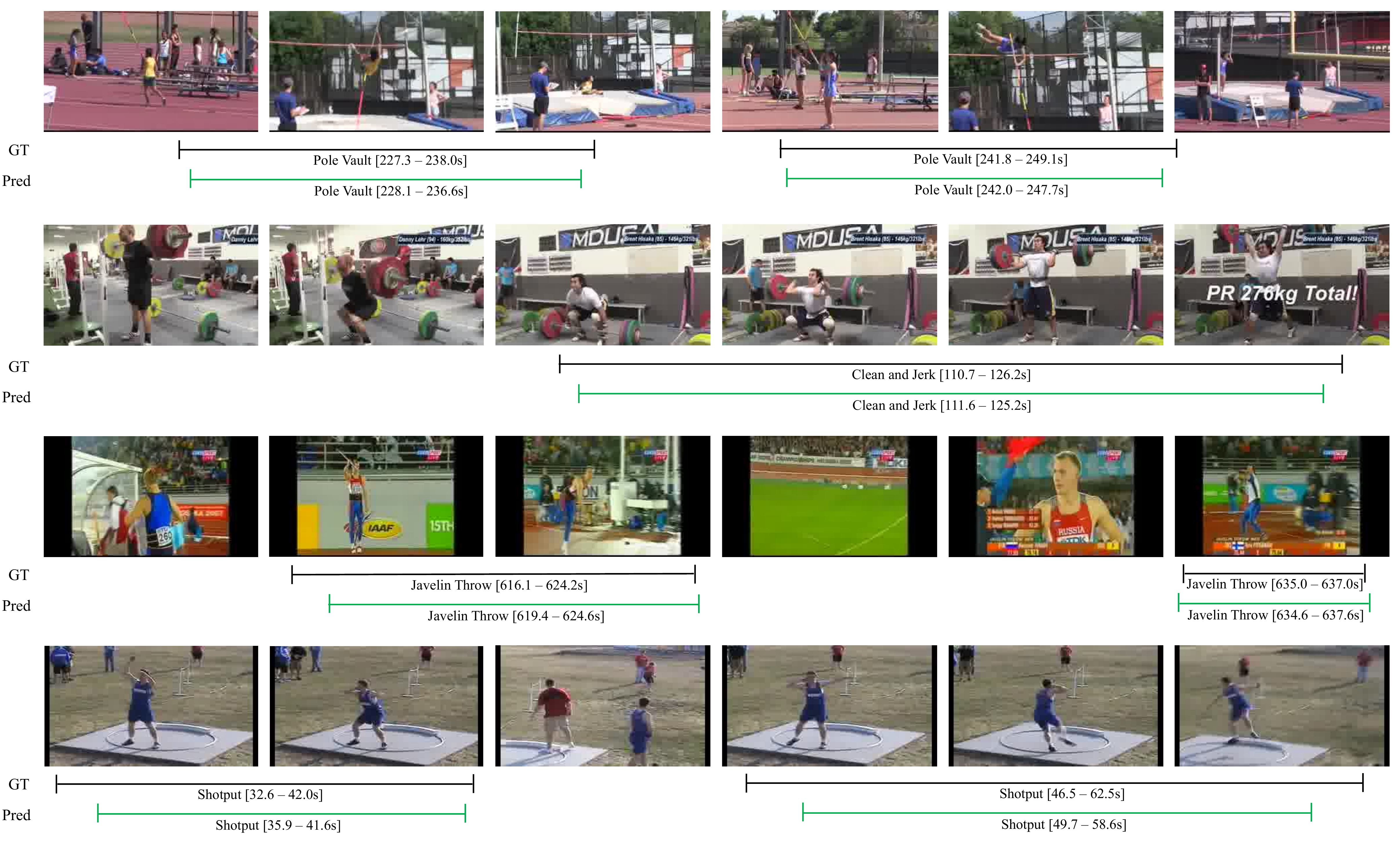}
\end{center}
   \caption{Qualitative visualization of the top detected activities by S$^3$D (best viewed in color) on four different activity categories in THUMOS'14 dataset: \textit{Pole Vault}, \textit{Clean and Jerk}, \textit{Javelin Throw} and \textit{Shotput}. Ground truth activity segments are marked in black and predicted activity segments are marked in green.} 
\label{fig:expresult}
\end{figure*}

\noindent \textbf{Qualitative Results.} We provide qualitative results to demonstrate the effectiveness and robustness of our proposed S$^3$D network. As shown in Figure~\ref{fig:expresult}, different video streams contain very diversified background context and different activity instances vary a lot in temporal location and scale. S$^3$D is able to predict the accurate temporal span as well as the correct activity category. Furthermore, S$^3$D can distinguish activity with minor differences such as the normal weightlifting compared to \textit{Clean and Jerk}. It is also capable of detecting the same activity sequence with different playing speed as shown in the \textit{Shotput} example. 

\noindent \textbf{Activity Detection Speed.} Since our model has a single-shot, end-to-end design with simple Conv3D building blocks, it is also very efficient. We benchmark our model on a GeForce GTX 1080 Ti GPU, and our S$^3$D can run much faster than real time at 1271 FPS. For comparison, previous top performing methods~\cite{shou2016temporal,DBLP:conf/mm/LinZS17,cdc_shou_cvpr17} have significantly lower FPS for the whole detection pipeline. Comparing to some recent works~\cite{Xu2017iccv,buch2017end} providing good runtime efficiency, our S$^3$D achieves much better accuracy.

\section{Conclusion}
	In this paper, we introduce S$^3$D, a Single Shot multi-Span Detector for temporal activity detection. We design a simple network architecture by  using only a fully Conv3D network on top of the raw video frames to jointly predict the temporal boundaries as well as activity categories. A key feature of S$^3$D is the use of multi-scale temporal span outputs attached to multiple temporal feature maps. With this framework, we achieved state-of-the-art performance on THUMOS'14 benchmark dataset, while being efficient to run much faster than real time on a single GPU.

\bibliography{egbib}

\begin{thebibliography}{43}
\providecommand{\natexlab}[1]{#1}
\providecommand{\url}[1]{\texttt{#1}}
\expandafter\ifx\csname urlstyle\endcsname\relax
  \providecommand{\doi}[1]{doi: #1}\else
  \providecommand{\doi}{doi: \begingroup \urlstyle{rm}\Url}\fi

\bibitem[Buch et~al.(2017{\natexlab{a}})Buch, Escorcia, Ghanem, Fei-Fei, and
  Niebles]{buch2017end}
S~Buch, V~Escorcia, B~Ghanem, L~Fei-Fei, and JC~Niebles.
\newblock End-to-end, single-stream temporal action detection in untrimmed
  videos.
\newblock In \emph{Proceedings of the British Machine Vision Conference
  (BMVC)}, 2017{\natexlab{a}}.

\bibitem[Buch et~al.(2017{\natexlab{b}})Buch, Escorcia, Shen, Ghanem, and
  Niebles]{buch2017sst}
Shyamal Buch, Victor Escorcia, Chuanqi Shen, Bernard Ghanem, and Juan~Carlos
  Niebles.
\newblock Sst: Single-stream temporal action proposals.
\newblock In \emph{Computer Vision and Pattern Recognition (CVPR), 2017 IEEE
  Conference on}, pages 6373--6382. IEEE, 2017{\natexlab{b}}.

\bibitem[Caba~Heilbron et~al.(2016)Caba~Heilbron, Carlos~Niebles, and
  Ghanem]{caba2016fast}
Fabian Caba~Heilbron, Juan Carlos~Niebles, and Bernard Ghanem.
\newblock Fast temporal activity proposals for efficient detection of human
  actions in untrimmed videos.
\newblock In \emph{Proceedings of the IEEE Conference on Computer Vision and
  Pattern Recognition}, pages 1914--1923, 2016.

\bibitem[Carreira and Zisserman(2017)]{carreira2017quo}
Joao Carreira and Andrew Zisserman.
\newblock Quo vadis, action recognition? a new model and the kinetics dataset.
\newblock In \emph{2017 IEEE Conference on Computer Vision and Pattern
  Recognition (CVPR)}, pages 4724--4733. IEEE, 2017.

\bibitem[Chen et~al.(2017{\natexlab{a}})Chen, Liao, Yuan, Yu, and
  Hua]{chen2017coherent}
Dongdong Chen, Jing Liao, Lu~Yuan, Nenghai Yu, and Gang Hua.
\newblock Coherent online video style transfer.
\newblock In \emph{Proc. Intl. Conf. Computer Vision (ICCV)},
  2017{\natexlab{a}}.

\bibitem[Chen et~al.(2017{\natexlab{b}})Chen, Yuan, Liao, Yu, and
  Hua]{chen2017stylebank}
Dongdong Chen, Lu~Yuan, Jing Liao, Nenghai Yu, and Gang Hua.
\newblock Stylebank: An explicit representation for neural image style
  transfer.
\newblock In \emph{Computer Vision and Pattern Recognition (CVPR), 2017 IEEE
  Conference on}, 2017{\natexlab{b}}.

\bibitem[Dai et~al.(2017{\natexlab{a}})Dai, Ng, and Davis]{8100129}
Xiyang Dai, Joe Yue-Hei Ng, and Larry~S Davis.
\newblock Fason: First and second order information fusion network for texture
  recognition.
\newblock In \emph{2017 IEEE Conference on Computer Vision and Pattern
  Recognition (CVPR)}, volume~00, pages 6100--6108, July 2017{\natexlab{a}}.

\bibitem[Dai et~al.(2017{\natexlab{b}})Dai, Singh, Zhang, Davis, and
  Qiu~Chen]{dai2017temporal}
Xiyang Dai, Bharat Singh, Guyue Zhang, Larry~S Davis, and Yan Qiu~Chen.
\newblock Temporal context network for activity localization in videos.
\newblock In \emph{Proceedings of the IEEE Conference on Computer Vision and
  Pattern Recognition}, pages 5793--5802, 2017{\natexlab{b}}.

\bibitem[Escorcia et~al.(2016)Escorcia, Heilbron, Niebles, and
  Ghanem]{escorcia2016daps}
Victor Escorcia, Fabian~Caba Heilbron, Juan~Carlos Niebles, and Bernard Ghanem.
\newblock Daps: Deep action proposals for action understanding.
\newblock In \emph{European Conference on Computer Vision}, pages 768--784.
  Springer, 2016.

\bibitem[Feichtenhofer et~al.(2016)Feichtenhofer, Pinz, and
  Zisserman]{feichtenhofer2016convolutional}
Christoph Feichtenhofer, Axel Pinz, and Andrew Zisserman.
\newblock Convolutional two-stream network fusion for video action recognition.
\newblock In \emph{Proceedings of the IEEE Conference on Computer Vision and
  Pattern Recognition}, pages 1933--1941, 2016.

\bibitem[Gaidon et~al.(2013)Gaidon, Harchaoui, and Schmid]{gaidon2013temporal}
Adrien Gaidon, Zaid Harchaoui, and Cordelia Schmid.
\newblock Temporal localization of actions with actoms.
\newblock \emph{IEEE transactions on pattern analysis and machine
  intelligence}, 35\penalty0 (11):\penalty0 2782--2795, 2013.

\bibitem[Girshick(2015)]{girshick2015fast}
Ross Girshick.
\newblock Fast r-cnn.
\newblock In \emph{Proceedings of the IEEE international conference on computer
  vision}, pages 1440--1448, 2015.

\bibitem[Girshick et~al.(2014)Girshick, Donahue, Darrell, and
  Malik]{girshick2014rich}
Ross Girshick, Jeff Donahue, Trevor Darrell, and Jitendra Malik.
\newblock Rich feature hierarchies for accurate object detection and semantic
  segmentation.
\newblock In \emph{Proceedings of the IEEE conference on computer vision and
  pattern recognition}, pages 580--587, 2014.

\bibitem[He et~al.(2016)He, Zhang, Ren, and Sun]{he2016deep}
Kaiming He, Xiangyu Zhang, Shaoqing Ren, and Jian Sun.
\newblock Deep residual learning for image recognition.
\newblock In \emph{Proceedings of the IEEE conference on computer vision and
  pattern recognition}, pages 770--778, 2016.

\bibitem[He et~al.(2018)He, Chen, Liao, Sander, and Yuan]{hmm2018deep}
Mingming He, Dongdong Chen, Jing Liao, Pedro~V Sander, and Lu~Yuan.
\newblock Deep exemplar-based colorization.
\newblock \emph{ACM Transactions on Graphics (Proc. of Siggraph 2018)}, 2018.

\bibitem[Jain et~al.(2014)Jain, Van~Gemert, J{\'e}gou, Bouthemy, and
  Snoek]{jain2014action}
Mihir Jain, Jan Van~Gemert, Herv{\'e} J{\'e}gou, Patrick Bouthemy, and Cees~GM
  Snoek.
\newblock Action localization with tubelets from motion.
\newblock In \emph{Proceedings of the IEEE conference on computer vision and
  pattern recognition}, pages 740--747, 2014.

\bibitem[J{\'e}gou et~al.(2010)J{\'e}gou, Douze, Schmid, and
  P{\'e}rez]{jegou2010aggregating}
Herv{\'e} J{\'e}gou, Matthijs Douze, Cordelia Schmid, and Patrick P{\'e}rez.
\newblock Aggregating local descriptors into a compact image representation.
\newblock In \emph{Computer Vision and Pattern Recognition (CVPR), 2010 IEEE
  Conference on}, pages 3304--3311. IEEE, 2010.

\bibitem[Jiang et~al.(2014)Jiang, Liu, Zamir, Toderici, Laptev, Shah, and
  Sukthankar]{jiang2014thumos}
YG~Jiang, J~Liu, A~Roshan Zamir, G~Toderici, I~Laptev, M~Shah, and
  R~Sukthankar.
\newblock Thumos challenge: Action recognition with a large number of classes,
  2014.

\bibitem[Lin et~al.(2017)Lin, Zhao, and Shou]{DBLP:conf/mm/LinZS17}
Tianwei Lin, Xu~Zhao, and Zheng Shou.
\newblock Single shot temporal action detection.
\newblock In \emph{Proceedings of the 2017 ACM on Multimedia Conference}, pages
  988--996. ACM, 2017.

\bibitem[Liu et~al.(2016)Liu, Anguelov, Erhan, Szegedy, Reed, Fu, and
  Berg]{liu2016ssd}
Wei Liu, Dragomir Anguelov, Dumitru Erhan, Christian Szegedy, Scott Reed,
  Cheng-Yang Fu, and Alexander~C Berg.
\newblock Ssd: Single shot multibox detector.
\newblock In \emph{European conference on computer vision}, pages 21--37.
  Springer, 2016.

\bibitem[Mettes et~al.(2015)Mettes, van Gemert, Cappallo, Mensink, and
  Snoek]{mettes2015bag}
Pascal Mettes, Jan~C van Gemert, Spencer Cappallo, Thomas Mensink, and Cees~GM
  Snoek.
\newblock Bag-of-fragments: Selecting and encoding video fragments for event
  detection and recounting.
\newblock In \emph{Proceedings of the 5th ACM on International Conference on
  Multimedia Retrieval}, pages 427--434. ACM, 2015.

\bibitem[Mettes et~al.(2016)Mettes, van Gemert, and Snoek]{mettes2016spot}
Pascal Mettes, Jan~C van Gemert, and Cees~GM Snoek.
\newblock Spot on: Action localization from pointly-supervised proposals.
\newblock In \emph{European Conference on Computer Vision}, pages 437--453.
  Springer, 2016.

\bibitem[Oneata et~al.(2013)Oneata, Verbeek, and Schmid]{oneata2013action}
Dan Oneata, Jakob Verbeek, and Cordelia Schmid.
\newblock Action and event recognition with fisher vectors on a compact feature
  set.
\newblock In \emph{Proceedings of the IEEE international conference on computer
  vision}, pages 1817--1824, 2013.

\bibitem[Oneata et~al.(2014)Oneata, Verbeek, and Schmid]{oneata2014lear}
Dan Oneata, Jakob Verbeek, and Cordelia Schmid.
\newblock The lear submission at thumos 2014.
\newblock 2014.

\bibitem[Perronnin et~al.(2010)Perronnin, S{\'a}nchez, and
  Mensink]{perronnin2010improving}
Florent Perronnin, Jorge S{\'a}nchez, and Thomas Mensink.
\newblock Improving the fisher kernel for large-scale image classification.
\newblock \emph{Computer Vision--ECCV 2010}, pages 143--156, 2010.

\bibitem[Ren et~al.(2015)Ren, He, Girshick, and Sun]{ren2015faster}
Shaoqing Ren, Kaiming He, Ross Girshick, and Jian Sun.
\newblock Faster r-cnn: Towards real-time object detection with region proposal
  networks.
\newblock In \emph{Advances in neural information processing systems}, pages
  91--99, 2015.

\bibitem[Shou et~al.(2016)Shou, Wang, and Chang]{shou2016temporal}
Zheng Shou, Dongang Wang, and Shih-Fu Chang.
\newblock Temporal action localization in untrimmed videos via multi-stage
  cnns.
\newblock In \emph{Proceedings of the IEEE Conference on Computer Vision and
  Pattern Recognition}, pages 1049--1058, 2016.

\bibitem[Shou et~al.(2017)Shou, Chan, Zareian, Miyazawa, and
  Chang]{cdc_shou_cvpr17}
Zheng Shou, Jonathan Chan, Alireza Zareian, Kazuyuki Miyazawa, and Shih-Fu
  Chang.
\newblock Cdc: Convolutional-de-convolutional networks for precise temporal
  action localization in untrimmed videos.
\newblock In \emph{CVPR}, 2017.

\bibitem[Simonyan and Zisserman(2014{\natexlab{a}})]{simonyan2014two}
Karen Simonyan and Andrew Zisserman.
\newblock Two-stream convolutional networks for action recognition in videos.
\newblock In \emph{Advances in neural information processing systems}, pages
  568--576, 2014{\natexlab{a}}.

\bibitem[Simonyan and Zisserman(2014{\natexlab{b}})]{simonyan2014very}
Karen Simonyan and Andrew Zisserman.
\newblock Very deep convolutional networks for large-scale image recognition.
\newblock \emph{arXiv preprint arXiv:1409.1556}, 2014{\natexlab{b}}.

\bibitem[Tang et~al.(2013)Tang, Yao, Fei-Fei, and Koller]{tang2013combining}
Kevin Tang, Bangpeng Yao, Li~Fei-Fei, and Daphne Koller.
\newblock Combining the right features for complex event recognition.
\newblock In \emph{Proceedings of the IEEE International Conference on Computer
  Vision}, pages 2696--2703, 2013.

\bibitem[Tran et~al.(2015)Tran, Bourdev, Fergus, Torresani, and
  Paluri]{tran2015learning}
Du~Tran, Lubomir Bourdev, Rob Fergus, Lorenzo Torresani, and Manohar Paluri.
\newblock Learning spatiotemporal features with 3d convolutional networks.
\newblock In \emph{Proceedings of the IEEE international conference on computer
  vision}, pages 4489--4497, 2015.

\bibitem[van Gemert et~al.(2015)van Gemert, Jain, Gati, Snoek,
  et~al.]{van2015apt}
Jan~C van Gemert, Mihir Jain, Ella Gati, Cees~GM Snoek, et~al.
\newblock Apt: Action localization proposals from dense trajectories.
\newblock 2015.

\bibitem[Wang and Schmid(2013)]{wang2013action}
Heng Wang and Cordelia Schmid.
\newblock Action recognition with improved trajectories.
\newblock In \emph{Proceedings of the IEEE international conference on computer
  vision}, pages 3551--3558, 2013.

\bibitem[Wang et~al.(2011)Wang, Kl{\"a}ser, Schmid, and Liu]{wang2011action}
Heng Wang, Alexander Kl{\"a}ser, Cordelia Schmid, and Cheng-Lin Liu.
\newblock Action recognition by dense trajectories.
\newblock In \emph{Computer Vision and Pattern Recognition (CVPR), 2011 IEEE
  Conference on}, pages 3169--3176. IEEE, 2011.

\bibitem[Wang et~al.(2014)Wang, Qiao, and Tang]{wang2014action}
Limin Wang, Yu~Qiao, and Xiaoou Tang.
\newblock Action recognition and detection by combining motion and appearance
  features.
\newblock \emph{THUMOS14 Action Recognition Challenge}, 1\penalty0
  (2):\penalty0 2, 2014.

\bibitem[Wang et~al.(2015)Wang, Xiong, Wang, and Qiao]{wang2015towards}
Limin Wang, Yuanjun Xiong, Zhe Wang, and Yu~Qiao.
\newblock Towards good practices for very deep two-stream convnets.
\newblock \emph{arXiv preprint arXiv:1507.02159}, 2015.

\bibitem[Wang et~al.(2016)Wang, Qiao, Tang, and Van~Gool]{wang2016actionness}
Limin Wang, Yu~Qiao, Xiaoou Tang, and Luc Van~Gool.
\newblock Actionness estimation using hybrid fully convolutional networks.
\newblock In \emph{Proceedings of the IEEE Conference on Computer Vision and
  Pattern Recognition}, pages 2708--2717, 2016.

\bibitem[Xu et~al.(2017)Xu, Das, and Saenko]{Xu2017iccv}
Huijuan Xu, Abir Das, and Kate Saenko.
\newblock R-c3d: Region convolutional 3d network for temporal activity
  detection.
\newblock In \emph{Proceedings of the International Conference on Computer
  Vision (ICCV)}, 2017.

\bibitem[Yeung et~al.(2016)Yeung, Russakovsky, Mori, and Fei-Fei]{yeung2016end}
Serena Yeung, Olga Russakovsky, Greg Mori, and Li~Fei-Fei.
\newblock End-to-end learning of action detection from frame glimpses in
  videos.
\newblock In \emph{Proceedings of the IEEE Conference on Computer Vision and
  Pattern Recognition}, pages 2678--2687, 2016.

\bibitem[Yu and Yuan(2015)]{yu2015fast}
Gang Yu and Junsong Yuan.
\newblock Fast action proposals for human action detection and search.
\newblock In \emph{Proceedings of the IEEE conference on computer vision and
  pattern recognition}, pages 1302--1311, 2015.

\bibitem[Zhang et~al.(2017)Zhang, Maei, Wang, and Wang]{zhang2017deep}
Da~Zhang, Hamid Maei, Xin Wang, and Yuan-Fang Wang.
\newblock Deep reinforcement learning for visual object tracking in videos.
\newblock \emph{arXiv preprint arXiv:1701.08936}, 2017.

\bibitem[Zhao et~al.(2017)Zhao, Xiong, Wang, Wu, Tang, and
  Lin]{zhao2017temporal}
Yue Zhao, Yuanjun Xiong, Limin Wang, Zhirong Wu, Xiaoou Tang, and Dahua Lin.
\newblock Temporal action detection with structured segment networks.
\newblock In \emph{The IEEE International Conference on Computer Vision
  (ICCV)}, volume~8, 2017.

\end{thebibliography}
\end{document}